\documentclass[sigchi]{acmart}
\AtBeginDocument{%
  \providecommand\BibTeX{{%
    \normalfont B\kern-0.5em{\scshape i\kern-0.25em b}\kern-0.8em\TeX}}}

\acmYear{2020}
\acmConference[IoT '20]{10th International Conference on the Internet of Things}{October 6--9, 2020}{Malmö, Sweden}
\acmBooktitle{10th International Conference on the Internet of Things (IoT '20), October 6--9, 2020, Malmö, Sweden}

\setcopyright{none}
\renewcommand\footnotetextcopyrightpermission[1]{}
\usepackage{xcolor}

\usepackage[normalem]{ulem}

\usepackage{siunitx} 

\definecolor{color_good}{RGB}{226,195,196}
\definecolor{color_fair}{RGB}{202,131,161}
\definecolor{color_poor}{RGB}{135,56,115}

\begin{document}

\title{Information-Driven Adaptive Sensing Based on Deep Reinforcement Learning}

\author{Abdulmajid Murad}
\affiliation{%
  \institution{Department of Information Security and Communication Technology}
    \institution{Norwegian University of Science and Technology, NTNU }
  \city{Trondheim}
  \country{Norway} }
\email{abdulmajid.a.murad@ntnu.no}

\author{Frank Alexander Kraemer}
\affiliation{%
  \institution{Department of Information Security and Communication Technology}
    \institution{Norwegian University of Science and Technology, NTNU}
  \city{Trondheim}
  \country{Norway} }

\author{Kerstin Bach}
\affiliation{%
  \institution{Department of Computer Science}
    \institution{Norwegian University of Science and Technology, NTNU}
  \city{Trondheim}
  \country{Norway} }

\author{Gavin Taylor}
\affiliation{%
  \institution{Department of Computer Science}
    \institution{United States Naval Academy}
  \city{Annapolis}
  \country{USA} }

\begin{abstract}
In order to make better use of deep reinforcement learning in the creation of sensing policies for resource-constrained IoT devices, we present and study a novel reward function based on the Fisher information value. This reward function enables IoT sensor devices to learn to spend available energy on measurements at otherwise unpredictable moments, while conserving energy at times when measurements would provide little new information.  This is a highly general approach, which allows for a wide range of use cases without significant human design effort or hyperparameter tuning. We illustrate the approach in a scenario of workplace noise monitoring, where results show that the learned behavior outperforms a uniform sampling strategy and comes close to a near-optimal oracle solution.
\end{abstract}



\maketitle

\section{Introduction}

 A fundamental use case in the Internet of Things (IoT) is to represent a phenomenon of the real world with an estimation or prediction model informed by measurements done by IoT sensing devices. %
 When the phenomenon is measured and predicted over time, the sensor devices therefore need to decide \emph{when} they should make a measurement. %
 More frequent measurements often allow for more accurate predictions; however,
 to spend their constrained resources as frugally as possible, sensor devices should avoid acquiring, processing, and transmitting measurements which are already accurately and confidently predicted by the representation model. %

 Generally, the optimal placement of measurements depends on the phenomenon to be observed. %
 In noise monitoring of working environments, for example, measurements during nights or holidays are less relevant and more predictable than those during office hours. %
 An optimal sensing strategy may also depend on the local environment and situation of each sensor individually, such as its energy budget. %
 Learning these strategies is a complex task, which may also vary with each use case and specific sensor instance, due to local differences. %
 We therefore see a need for more autonomy through learning, so sensor devices can exhibit appropriate behavior for their specific situation without human guidance ~\cite{chincoli2018self, udenze2009indirect,fraternali2019aces, khan2014online}.%

 Reinforcement learning (RL) is one way to achieve such individual autonomous behavior. %
 It has proven suitable for sequential decision-making under uncertainty, and has been applied in IoT and wireless sensor networks for a variety of problems~\cite{Aoudia2018,Zhou2016,Fraternali2018,Hsu2009a,Shresthamali2017,murad2019autonomous}. %
 Current approaches, however, focus on generic performance measures, such as maximum duty cycle or optimal network traffic. %
 Such metrics can be useful if they are well correlated with the goals the system should achieve, but such correlation is not guaranteed, and their optimization may diverge from optimizing the true goals of the system. %

 Instead, we consider a system in which a continuous prediction is made of a phenomenon only sporadically measured by the IoT device in its entirety. %
 The goal is not to measure as frequently as possible, as in optimizing duty cycle or network activity; the goal is to make the prediction as accurate as possible. %
 Therefore, we want to make the IoT device learn how to schedule measurements so that they contribute in times when the prediction is unconfident or inaccurate, contributing optimally to the quality of the overall prediction. %

 For that, we define a novel reward function inspired by the work of Kho et al.~\cite{kho2009decentralized} that is based on the Fisher Information of the observed phenomena. %
 This constitutes a novel and general approach for the application of RL in IoT that can be applied to a wide variety of use cases. %
 Results show that the learned policies outperform uniform sampling for a case study with regard to model quality, and come close to oracle solutions. %

 In Sect.~\ref{sec:related}, we will outline related work, before defining our system setup and problem in Sect.~\ref{sec:system}. %
 We then present our prediction model and evaluation criteria in Sect.~\ref{sec:gp}, which is the basis for the frugal operation policy presented in Sect.~\ref{sec:frugal}. %
 We close with an evaluation of our results and a discussion.

\section{Related Work}\label{sec:related}

Adaptive sampling addresses the problem of \emph{when}, \emph{where}, or \emph{what} to sample, subject to some sampling budget constraints. %
This corresponds to, respectively, monitoring temporally-varying, spatially-varying, or multiple data types-varying environmental fields. %
Within the context of adaptive sampling, a large number of prior works have applied RL for power management of constrained, wireless sensors. %
In energy-harvesting sensors, one objective may be to increase the number of samples while maintaining perpetual operation ~\cite{Zhou2016,Fraternali2018,Hsu2009a,Shresthamali2017}. %
These works use tabular methods, such as SARSA or Q-learning, in which the state and action spaces are discrete and small enough to express the value function in tabular form. %
Other works~\cite{Aoudia2018, Ortiz2017} use linear functions to approximate power-managing policies. %
Most of the previous works include energy directly in the reward function, while work in~\cite{murad2019autonomous} proposes a utilitarian reward function that maximizes a system's utility based on the achieved duty cycle.%

With respect to battery-powered sensors, the main objective is to minimize energy consumption by reducing the number of samples while maintaining an acceptable level of service. %
For example, Dias et al.~\cite{dias2016} use a Q-learning algorithm to learn an adaptive sampling policy to minimize power consumption while not missing environmental changes. %
They define the action space to be a range of possible sampling intervals, and the reward function to be proportional to the amount of energy saved, subject to a constraint that the difference between consecutive measurements is less than a threshold value. %
Cobb et al.~\cite{cobb2016adaptive} apply RL to learn a sampling policy on accelerometer data on lions and show that it is feasible to achieve a reconstruction accuracy of 51 \% with 73\% reduction in energy consumption. %
They propose a reward function based on the sampling rate and data variance, then test tabular Q-learning, Deep Q-learning with NN, and Deep Q-learning with LSTM. %
They also present a heuristic sampling algorithm that outperforms these RL methods. However, they argue that using a heuristic algorithm would be restrictive and not very adaptable.

We observe an increasing attention in IoT towards incorporating prediction models with adaptive sampling, i.e., model-driven adaptive sampling. %
Ling et al. ~\cite{ling2016gaussian} address adaptive sampling when predicting an environmental field by presenting an adaptive GP planning framework. %
They integrate planning and learning by framing the problem as a Bayesian sequential decision with a value function.  %
However, they do not derive an exact planning policy due to the uncountable set of candidate measurements, and thus large possible sequences of posterior GPs.  %
Instead, they use a Lipschitz-continuous reward function to derive an asymptotically-optimal policy. %
They evaluate their framework, among others, on a simulated energy-harvesting task in which a rover harvests wind energy while exploring a polar region. %
Muttreja et al.~\cite{muttreja2006active} model sensor data using sparse Gaussian processes and propose an active-learning based sampling algorithm. %
The objective of the sampling algorithm is to maintain the model confidence within pre-specified bounds while minimizing energy consumption. %
Similarly, Monteiro et al.~\cite{monteiro2017dpcas} combine a data prediction model with adaptive sampling, where they build a model based on an extension of Holt's Method. %
The algorithm greedily samples until it keeps the prediction error under a threshold value. %
Most similar to our work are those that use Gaussian processes in each sensor for prediction models and Fisher information as a basis for the sampling algorithms~\cite{kho2009decentralized, osborne2012real}. %
In contrast to our work, where we use a learned sampling policy, these works use manually fine-tuned sampling algorithms. %
 
In a somewhat different setting, some other works use variance-based adaptive sampling without prediction models~\cite{laiymani2013adaptive, silva2017litesense, harb2017energy}. %
They manually designed sampling algorithms that adjust the sampling rate according to the variation in the environment. %
Salim et al. ~\cite{salim2016adaptive} propose an algorithm that uses analysis of variance with Fisher test to adapt the sensing frequency according to the environmental variation. %

\section{System Setup}\label{sec:system}
The system in our case study measures the noise levels in a working environment, and we study the operation of a single conceptual sensor device. %
The data that is the basis for the case study was collected using a commercial system and is further detailed in ~\cite{kraemer2019energy}. %
The noise levels are aggregated as so-called \emph{equivalent continuous sound levels} $L_{\mathit{Aeq}}$, which is a standard indicator used for noise measurement ~\cite{ISO2016}. %
For this case study, we select time slots with a length of 15 minutes. %
This seems an appropriate resolution for use cases where the working environment should be evaluated. %
This means there are 96-time slots during a day, or 672 per week.

Often, we are not just interested in measuring a phenomenon, such as the noise levels here. %
We rather want a model of the phenomena which can predict or estimate noise levels based on explanatory variables, even when the IoT device has not recently made measurements, providing a service as close as possible to a continually-monitoring sensor. %
For many application use cases, such predictive power is more useful than only retrospective logging: %
A noise monitoring application can for instance recommend silent workplaces, or detect significant outliers in noise corresponding to what is usual. %
A forecast model implies uncertainty; we do not expect it to represent only points of actual observations, but estimated values that correspond to the actual ones with some certainty. %
This different view on an IoT sensing system allows us to optimize the behavior of a sensing device. %
When its energy budget is constrained so that it can only sample a subset of the time slots, the sensor device must be strategic about which time slots to observe. %
The hypothesis---which we confirm in this paper---is that once the sensor device selects the time slots to sample wisely, it can produce prediction models with high accuracy and high certainty with only a subset of possible observations. %
Intuitively speaking, the sensor should only spend energy on observations that are ``interesting'' with respect to the prediction model, where the prediction is uninformed or unconfident. %
The challenge is developing such a sampling policy that can anticipate time slots worth covering. %

\begin{figure}[tb]
   \centering
   \includegraphics[width=\linewidth]{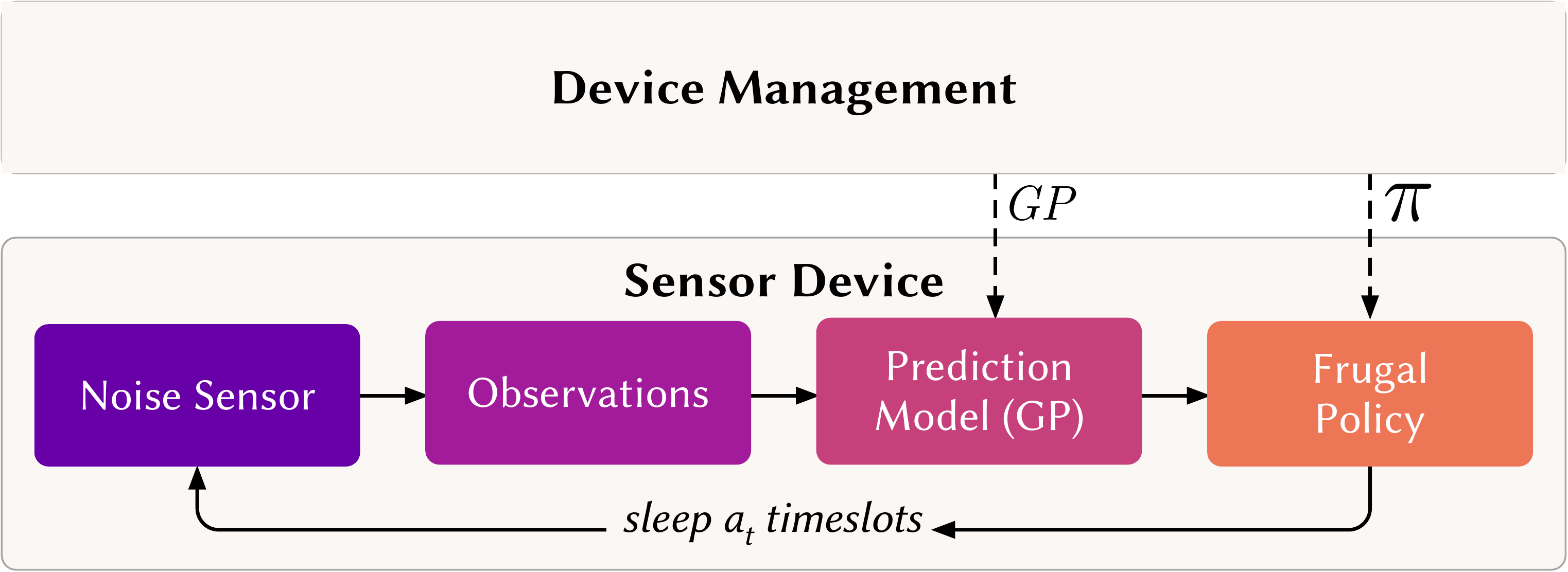}
   \caption{Overview of the system and the sensor operation.}
   \label{fig:overview}
 \end{figure}

An overview of our process is illustrated in Fig.~\ref{fig:overview}.\footnote{Data and code is available at \href{https://github.com/Abdulmajid-Murad/adaptive-sensing}{https://github.com/Abdulmajid-Murad/adaptive-sensing}} %
The sensor takes a noise measurement during a 15-minute time slot and stores the observed value. %
As a prediction model, we use a Gaussian process (GP) with explanatory
variables as inputs. 
Given the explanatory variables and the output of the Gaussian process, the sensor then chooses the number of time steps to sleep before the next measurement; this decision-making is the frugal policy $\pi$. %
It outputs $a_t=1$ to observe the immediate next time slot, or $a_t=2$ to hop over one-time slot, and so on. The higher $a_t$, the longer the sensor device can stay in sleep mode. %
After the sensor wakes up, it makes a measurement, and the GP is updated with the latest observation to make a new prediction.

The policy $\pi$ is the result of reinforcement learning, encoded as a neural network. %
We will detail this step in Sect. ~\ref{sec:drl}. %
Training a reinforcement learning agent is a computationally intensive phase and hence best executed on a server with sufficient computational resources, for instance, as part of the device management. %
The training step requires training data as input, both for the reinforcement step that creates the policy and to train an initial version of the Gaussian process as a prediction model. %
For that, we use two weeks of data that the sensor collected uniform sampling before going into frugal mode. %

To be explicit, a sample during the frugal policy has the following three purposes:
\begin{itemize}
  \item The data point provides an entirely accurate measurement for that time period.
  \item The data point provides a training point for the refinement of the Gaussian process for future predictions.
  \item The data point provides a training point for the refinement of the RL policy.
\end{itemize}

In the following, we have a close look at the Gaussian process as a prediction model, the use of RL for building the frugal policy and especially the design of the reward function.%

\section{Gaussian Processes as Predictors}\label{sec:gp}

We use a Gaussian process (GP) as a representation model of the monitored phenomenon~\cite{williams2006gaussian}.
The GP produces a Bayesian probabilistic model of the monitored phenomenon over a period of time by predicting a Gaussian distribution over possible measurements with expected value $\mu_*$ and variance $\pm \sigma_*^2$ at every time slot.

\begin{figure}[t]
  \centering
  \includegraphics[width=\linewidth]{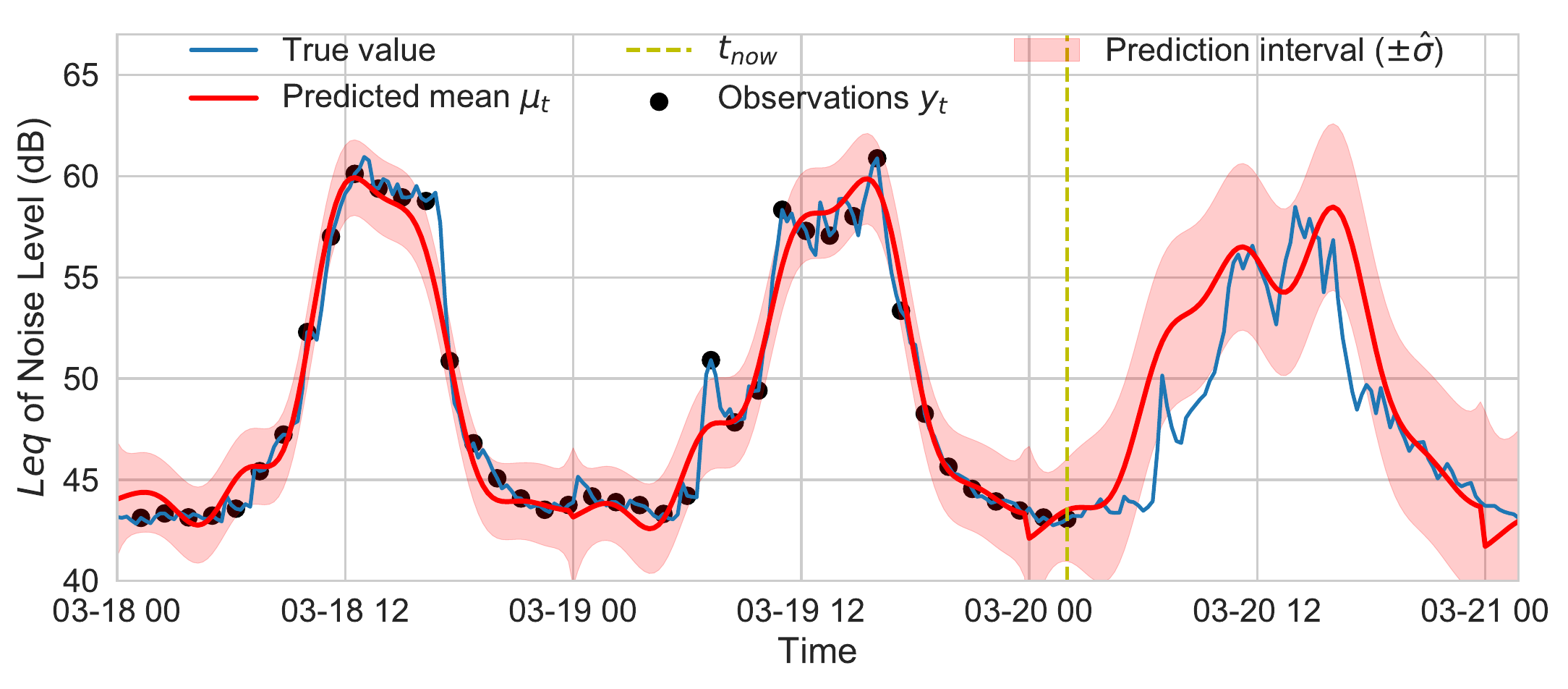}
  \caption{Gaussian Process regression applied to the noise monitoring and performing short-term prediction}
  \label{fig:gp1}
\end{figure}

Figure~\ref{fig:gp1} shows the estimation of noise levels of a GP over several days.
The blue line shows the true value, 
the red line shows the mean of the GP's estimation, and the red shading indicates the standard deviation of the GP's estimation.
The GP is trained on the observations of the true value marked with filled circles until $t_{\mathit{now}}$.
Until that time, the GP's estimation works as an interpolation between the observations, deviating from occasional outliers.
Beyond $t_{\mathit{now}}$, the GP's estimation works as a forecast.
As the uncertainty of the GP increases, the variance becomes larger after $t_{\mathit{now}}$.

Before using the GP for estimation, it must be trained with a training dataset $\mathbf{\mathcal{D}} =\{\mathbf{X}, \mathbf{y }\} $, which consists of measurement collection  $ \mathbf{y} = [\begin{matrix} y_1 & y_2  & \dots  & y_N \end{matrix}]^T$, and their corresponding inputs $ \mathbf{X} = [\begin{matrix} \mathbf{x}_1 & \mathbf{x}_2  & \dots  & \mathbf{x}_N \end{matrix}]^T$.  
Every entry of the input $\mathbf{x}_k $ is a feature vector that consists of measurement time and other explanatory variables ~\cite{kraemer2019energy}.
The GP produces a prediction distribution based on similarity with training examples, where similarity is defined by the kernel function. 
In our case, we chose the commonly used Matern kernel function, which is a generalization of the radial basis function (RBF kernel):
\begin{equation}
\mathbf{K}_{mat}(\mathbf{x}_i, \mathbf{x}_j) = \sigma_{mat}^2 \frac{2^{1-v}}{\Gamma (\nu)} \left(  \sqrt{2\nu}  \frac{\mathbf{d}}{\rho}  \right)^{\nu} \mathbf{K}_{\nu}   \left(  \sqrt{2\nu}  \frac{\mathbf{d}}{\rho}  \right)
 \label{eq:matern_kernel}
\end{equation}
Where $\mathbf{d} = ||\mathbf{x}_i - \mathbf{x}_j||$ is the Euclidean distance between the two measurement points $ \mathbf{x}_i$ and $\mathbf{x}_j$.
The kernel's parameters $\sigma^2$, $\rho$, and $\nu$ can be hand-tuned or estimated by maximizing the likelihood of the observations in the training dataset. 
$\Gamma$ is the gamma function, and $\mathbf{K}_{\nu} $ is the Bessel function. 
Due to its finite differentiability, the Matern kernel is better able to capture a phenomenon than RBF. 
The Bessel and gamma functions are derived from the Fourier transform of a finite positive measure. 
The measure is a stochastic differential equation of Laplace type that works well in more than one dimension. 

Additionally, we added a periodic kernel function to model the periodicity in the environmental variable.
\begin{equation}
\mathbf{K}_{per} (\mathbf{x}_i, \mathbf{x}_j) = \sigma_{per}^2 \exp \left (  -\frac {2 \sin^2(\frac{ \pi \mathbf{d}}{p})}{\ell^2} \right)
 \label{eq:periodic_kernel}
\end{equation}
Where $\ell$ is the length scale, $\sigma^2$ is the variance, and $p$ is the period of the kernel.  
Again, these parameters can be hand-tuned or estimated by maximizing the likelihood. 
The resulting overall kernel is a combination of the Matern and the periodic kernels.
\begin{equation}
 \mathbf{K}(\mathbf{x}_i, \mathbf{x}_j) = \mathbf{K}_{mat}(\mathbf{x}_i, \mathbf{x}_j) + \mathbf{K}_{per} (\mathbf{x}_i, \mathbf{x}_j)
 \label{eq:overall_kernel}
\end{equation}

\subsection{Evaluating Prediction Models}
\label{sec:evaluating_prediction_models}

\begin{figure}[tb]
  \centering
  \includegraphics[width=\linewidth]{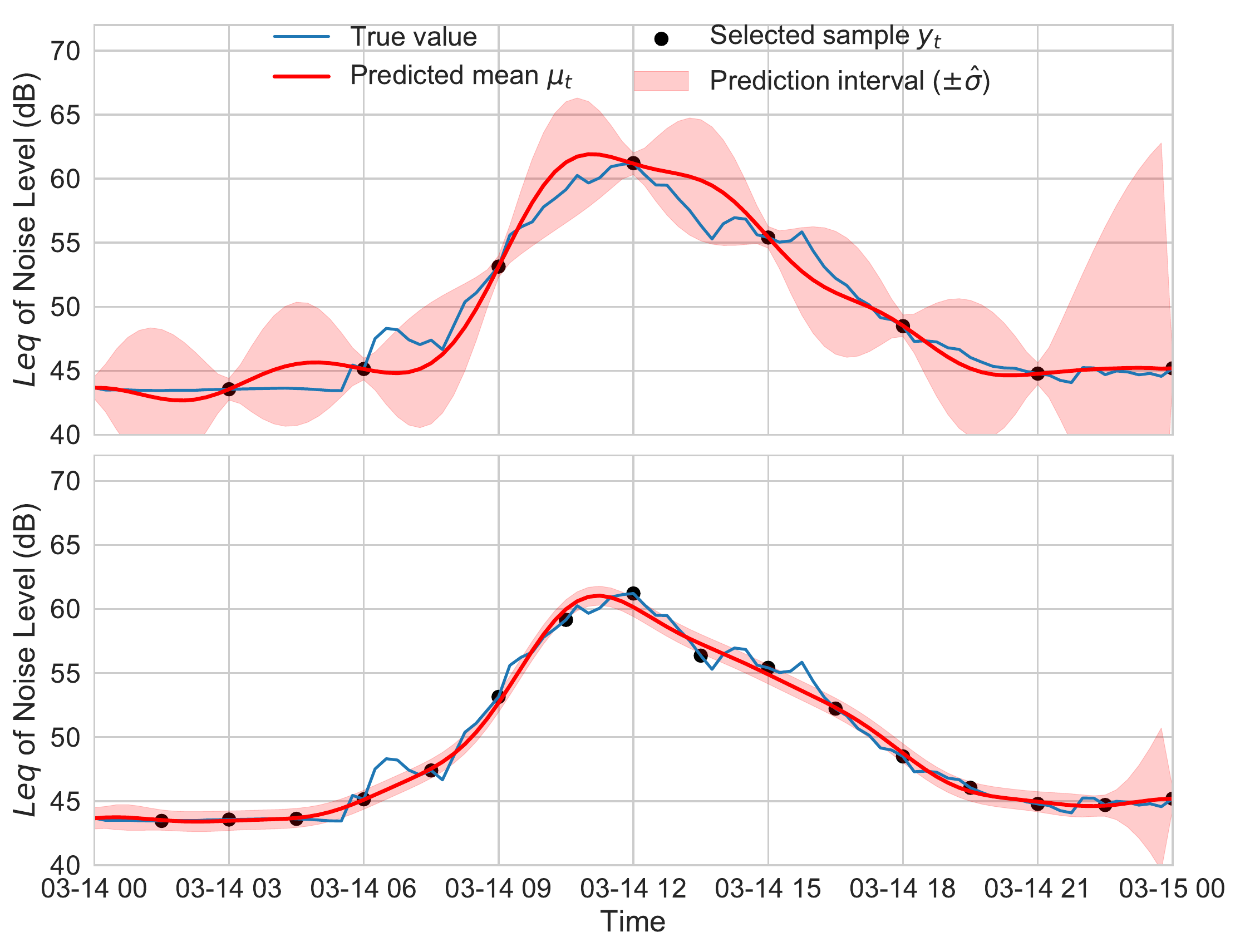}
  \caption{Comparison of two different predictive models. The top model has a less certain predictions with FI = 167.84, and RMSE = 1.42. The bottom model has more certain prediction with FI = 1111.73, and RMSE = 0.70.}
  \label{fig:fi}
\end{figure}

To assess the quality of the representation model generated by a GP, we can use the root mean square error ($\mathit{RMSE}$) between the predicted mean  $\mu_t$ and the true values $y_t$ over a period of time $T$.
\begin{equation}
	\mathit{RMSE_T} = \sqrt{\frac{1}{T} \sum\limits_{T} (\mu_t - y_t)^2}
	\label{eq:rmse}
\end{equation}
However, the $\mathit{RMSE}$ metric overlooks an important feature of a Bayesian probabilistic model, which is its prediction interval $\pm \sigma_t^2$. In other words, a prediction should be both accurate and confident. %
Therefore, we want a metric that also quantifies the uncertainty represented by the model. %
Figure~\ref{fig:fi} shows two different GP predictive distributions of noise levels over one day. %
The lower model was built with more observations, resulting in more certain predictions with smaller uncertainty $\sigma_t^2$ (or higher precision $\frac{1}{\sigma_t^2}$). %
As a metric for the certainty of a model over a time period $T$ we can take the precision mean: %
\begin{equation}
    \mathit{FI}=\frac{1}{T}\sum\limits_{T} \frac{1}{\sigma_t^2}
       \label{eq:fi}
\end{equation}
This metric is called \emph{Fisher information}, and has been used in many previous works in the field WSN for data fusion, sensor selection, or adaptive sensing~\cite{rogers2006computational,kho2009decentralized,wang2016dynamic, yilmaz2013sequential}. %

\section{Frugal Operation Mode}\label{sec:frugal}

In the frugal operation mode, the sensor device only has a limited energy budget over a period $T$. %
We assume that the sensor can make no more than $N$ observations in $T$, and each observation costs a constant amount of energy.
By executing a policy, the sensor device decides which time slots will be sampled. %

\begin{figure}[tb]
  \centering
\includegraphics[width=\linewidth]{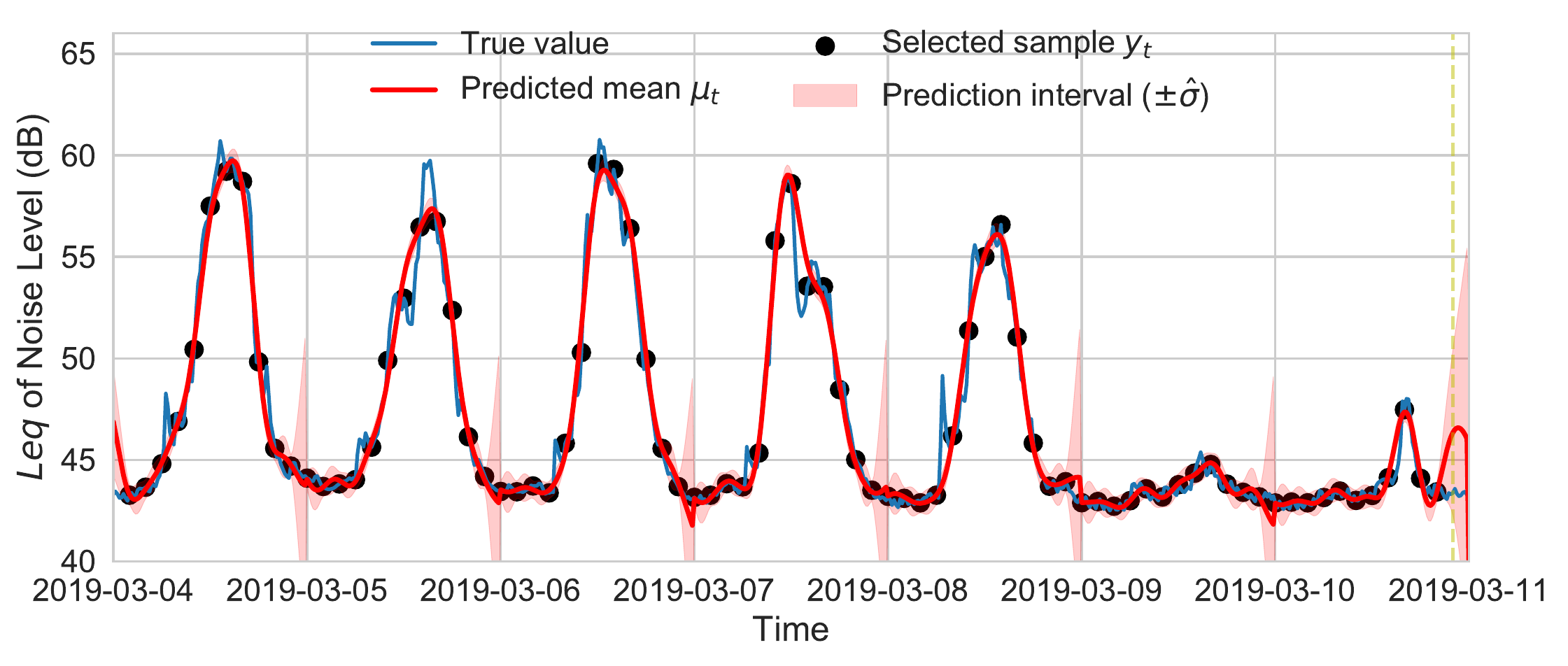}
  \caption{GP model updated with uniform non-adaptive sampling, having FI = 4826.50, and RMSE = 0.94}
  \label{fig:uniform_sampling}
\end{figure}

A straightforward sampling policy is uniform sampling, in which a time period $T$ is divided by $N$ equidistant measurements.
Figure ~\ref{fig:uniform_sampling} shows a GP prediction model fitted from 100 observations distributed uniformly over a one-week period. %
It results in an RMSE of 0.94 and an FI of 4826.50. This approach provides a steady flow of information, but may result in samples being taken where the prediction is already confident and accurate, and therefore sufficient without the sample. %

\begin{figure}[tb]
  \centering
  \includegraphics[width=\linewidth]{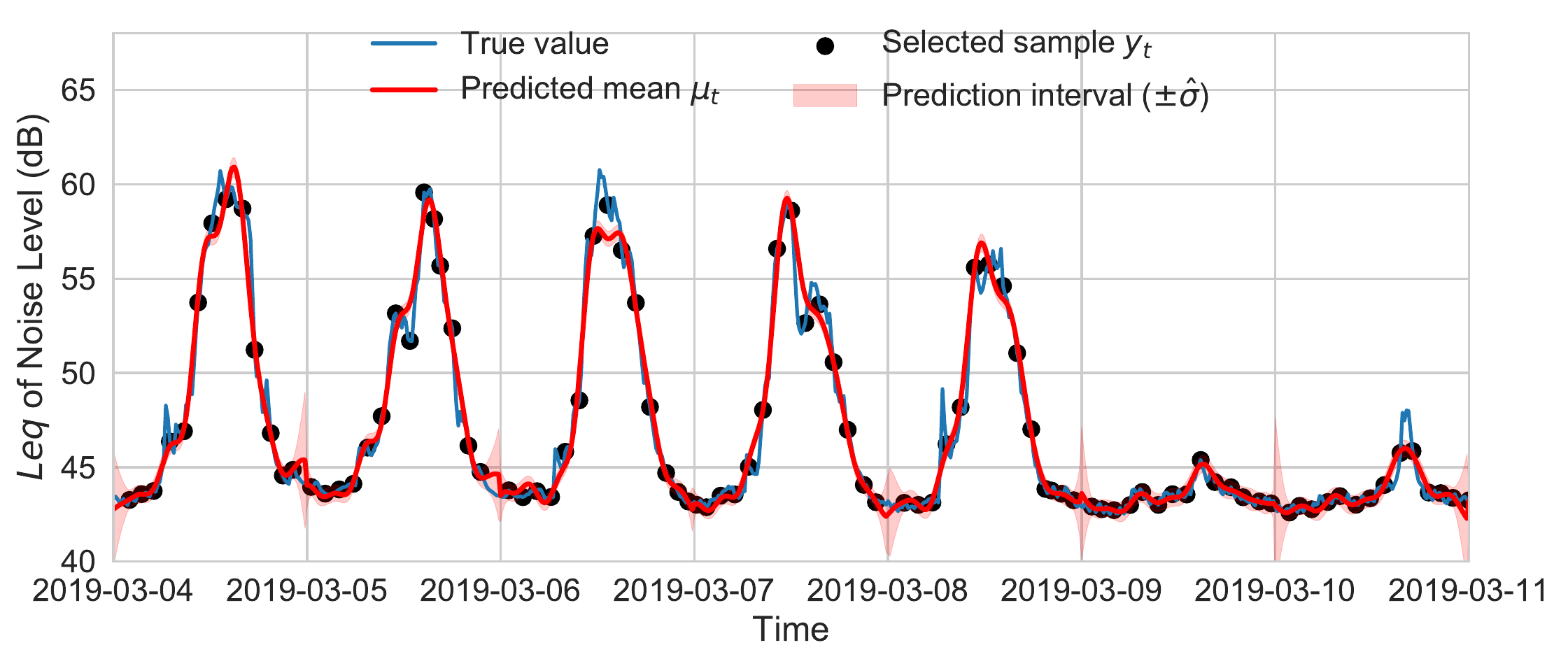}
  \caption{GP model updated with an offline oracle sampling, having FI = 7331.03, and RMSE = 0.85}
  \label{fig:greedy}
\end{figure}

To gain an understanding of the potential optimal sampling, Kho et al.~\cite{kho2009decentralized} describe \emph{greedy optimal adaptive sampling}. %
It is an offline oracle solution that requires knowledge of future observations and is hence not usable during operation, but establishes a near-optimal baseline. %
(The authors also outline an optimal solution, which is computationally unfeasible, even offline.) %
Greedy optimal adaptive sampling greedily allocates samples based on the mean FI gain over a day. %
Figure~\ref{fig:greedy} shows the same GP prediction model, again fitted with 100 observations. %
When trained with observations with this oracle policy, the GP's RMSE is reduced to 0.85 compared to that of one trained with samples from the uniform policy; additionally, the uncertainty is reduced, i.e., the predictor has a higher FI of 7331.03. %
This is a considerable improvement over the uniform policy.

\subsection{Deep Reinforcement Learning Policies}\label{sec:drl}

The sampling policies described above outline the task for a reinforcement learning agent in the system: it should learn a sampling policy that is close to the optimal one, but without using knowledge of the future.

\begin{figure}[t]
  \centering
  \includegraphics[width=\linewidth]{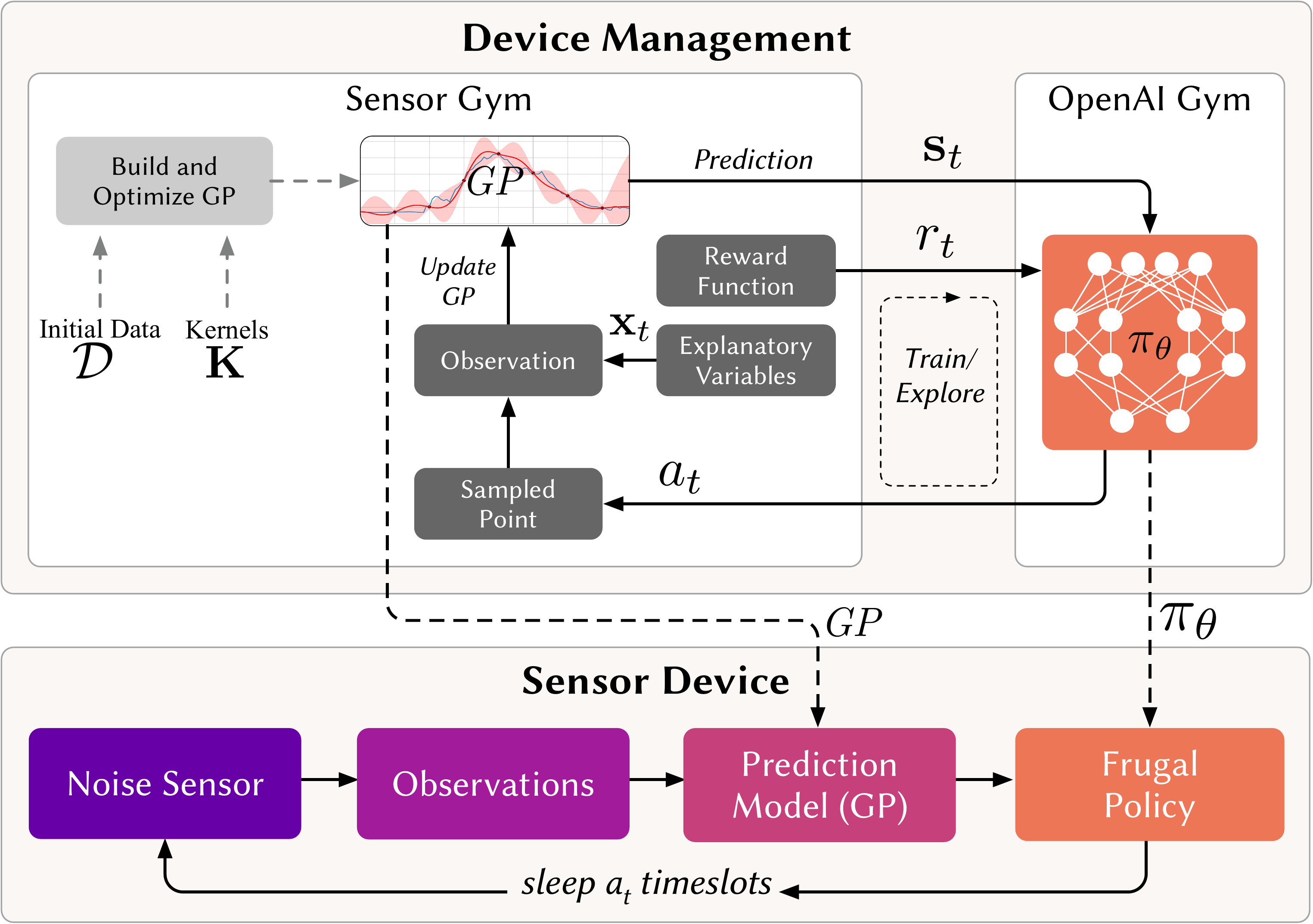}
  \caption{Architecture for training a model-driven sampling policy with reinforcement learning}
  \label{fig:rl_training}
\end{figure}

Figure ~\ref{fig:rl_training} illustrates the process of training a reinforcement learning agent to learn an adaptive sampling task, which is based on the IoT Sensor Gym~\cite{murad2019iot}. %
Generally, a reinforcement learning task is modeled as a Markov decision process $M=(\mathcal{S},\mathcal{A},\mathcal{P},R,\gamma)$, where $\mathcal{S}$ is the set of the problem's states; $\mathcal{A}$ is the set of actions an agent can take; $\mathcal{P}$  is the states' transition probability; $R:\mathcal{S}\rightarrow\Re$ is the reward function; and $\gamma\in (0,1)$ is a discount factor. %
In our case study, at each time $t$, the state ($s_t \in \mathcal{S}$) consists of: the GP's future prediction $(\mathbf{\mu}_t^h, \mathbf{\sigma}_t^h)$ over a prediction horizon $h$; the battery level $B_t$; and the explanatory variables $\mathbf{x}_t^h$, which consist of weekday, time of day, and whether a given day is a holiday.
\begin{equation}
	s_t = [\begin{matrix}  \mathbf{\mu}_t^h & \mathbf{\sigma}_t^h   & B_t  & \mathbf{x}_t^h  \end{matrix}]
 	\label{eq:state_space}
\end{equation}
In response to receiving the state $s_t$, the agent takes an action ($a_t \in \mathcal{A}=(1, h)$) which is the index for the next time step to make a sample, i.e., sleep $a_t$ time slots. %
The environment, then, transits to a new state according to the transition probability ($p(s'|s,a)~ \forall s,s'\in\mathcal{S},a\in\mathcal{A}$), and the agent receives a reward $r_t=R(s_t)$ according to the reward function $R$, explained later in detail. %
By interacting with the environment, the agent learns a policy $\pi:\mathcal{S}\rightarrow\mathcal{A}$, which can be evaluated using the value function of a state $V^\pi(s)$: 
\begin{align}
  \label{eqn:value}
  V^\pi(s)=&R(s)+\int_\mathcal{S} p(s'|s,\pi(s))V^\pi(s')ds'
\end{align}
By learning, the agent tries to find a behavior policy that maximizes the value for all states; there exist a variety of learning algorithms in RL ~\cite{sutton2011reinforcement}, which intend to produce such policies. %
In this paper, we use a policy gradient algorithm, specifically Proximal Policy Optimization (PPO) ~\cite{Schulman2017}.
In PPO, policies ($\mathbf{\pi_{\theta}}$) are represented by deep neural networks, parameterized by $\mathbf{\theta}$, which can produce actions for any state from a continuous state space. Compared to tabular methods such as SARSA or Q-learning, which are sometimes seen in literature applying RL to IoT, PPO eliminates the need for manual state-space discretization, allowing for better generalization and potentially less hand-tuning by the user.

\subsection{Information-Based Reward Function}
A key requirement for a successful application of reinforcement learning is to design a reward function that frames the goal of an application and guides the learning towards a desirable behavior.  %
However, many goals are difficult to translate into scalar values or are intractable to learn.  %
To overcome this difficulty, many of the previous works use reward shaping, which is a way of incorporating domain knowledge by rewarding easily quantified subgoals to accelerate learning, such as energy-based reward~\cite{dias2016, Hsu2009a, Ortiz2017, Shresthamali2017}.  %
Although effective in specific domains, energy-based rewards can be restrictive and may lead to undesired emergent behavior in new domains, such as high variance in duty cycle which is not appropriate for IoT nodes. 
This is because energy is only a resource to be managed, not a goal to be optimized.  %
In our previous work~\cite{murad2019autonomous}, we addressed this issue by introducing a utilitarian reward function that maximizes the utility of a system by rewarding it for high duty cycles. %
This means, a system gets more reward the more samples it manages to take, no matter if these samples are valuable or not. %

The reward functions outlined above only utilize metrics that are more or less correlated with the application goal, but not identical to it. %
In our setting, the system's goal is to construct a model that represents the underlying phenomenon as closely as possible. %
Since the sensor has a limited sampling budget, and not all samples are valuable, the objective of the reinforcement learning agent is to select the most informative observations concerning the representation model. %
Therefore, we design a new reward function that reflects this goal, i.e., rewarding the agent based on the information content of its selected observations. %
By doing so, we judge the agent's performance by the metric that we are trying to maximize, and there is a good alignment between maximizing reward and the true goal of the system. %
Our hypothesis is that this leads to a better performance of the device, as it will learn to spend its constrained energy budget where it has the best-expected effect. %
Going back to Sect.~\ref{sec:evaluating_prediction_models} where we defined the metric for the prediction model, we see that the proposed metric can also work as a reward function. %
At the end of a period of time $T$, a day in our case, the RL agent receives a reward based on the quality of the representation model, which is fitted by observations selected by the agent. 
The quality of the model is abstracted by the mean Fisher information over $T$: 
\begin{equation}
R(\mathbf{s}_T)= \frac{1}{T}\sum\limits_{T} \frac{1}{\sigma_t^2}
 \label{eq:reward_function}
\end{equation}
where $\frac{1}{\sigma_t^2}$ is the prediction confidence at point $t$.
Accordingly, we propose to use the Fisher information directly as a reward function. %
Fisher information does not depend on unobserved samples as compared to the RMSE,
and decodes the application goal into a scalar value that is easy to learn. %
Hence, it eliminates the need for reward shaping, and it can be applied to large, scalable applications without requiring a system-specific knowledge. %

\section{Evaluation Results}\label{sec:evaluation}
\begin{figure}[t]
  \centering
  \includegraphics[width=\linewidth]{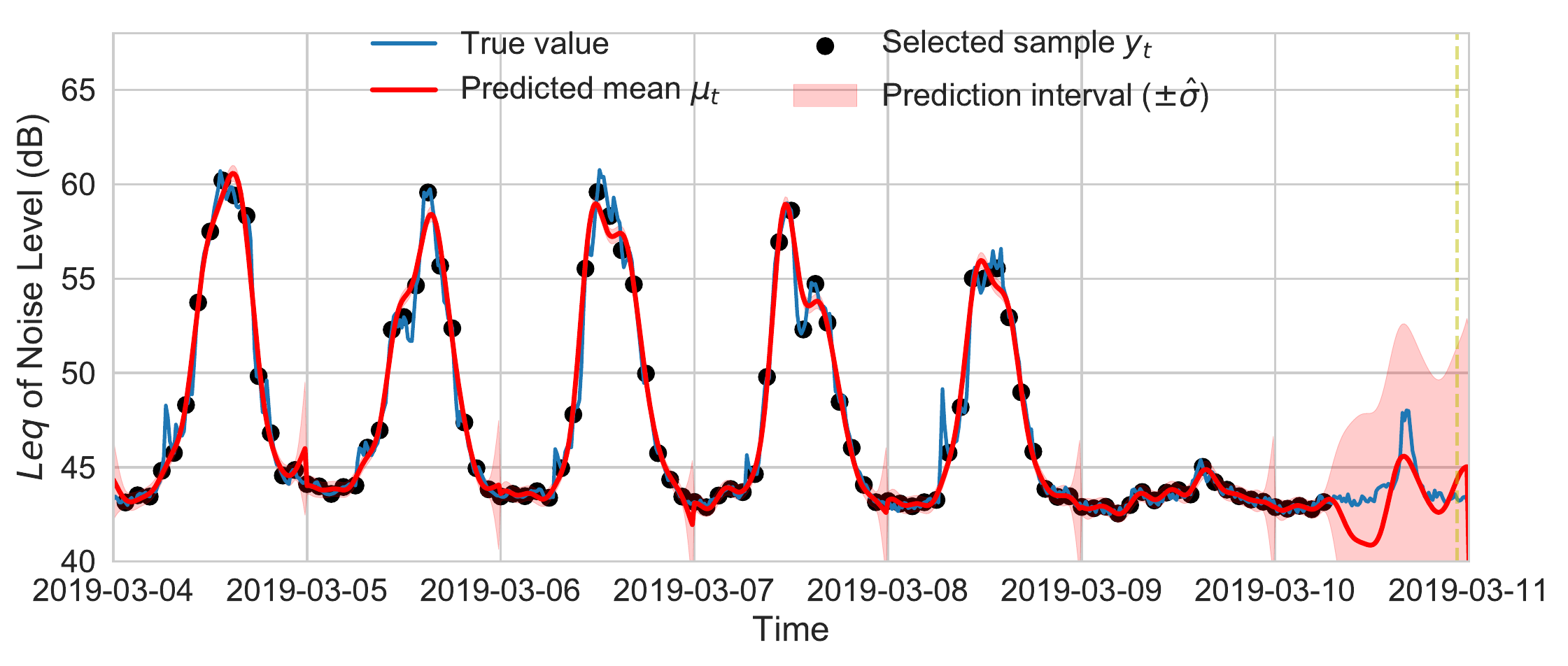}
  \caption{GP model fitted with observations selected using an RL sampling policy, collecting information value of FI = 7274.84, and RMSE = 0.86}
  \label{fig:RL_sampling}
\end{figure}

After training approximately 120 RL agents with different hyperparameters, we tested their performance over a different one-week period.
Figure ~\ref{fig:RL_sampling} shows a GP prediction model with observation selected using an RL agent. 
The GP trained on data chosen using an RL sampling policy achieved a better
information gain (FI = 7274.84) and a lower RMSE  (RMSE = 0.86) compared to the
uniform sampling, and approaches the near-optimal policy implemented by
the oracle.

\begin{figure}[t]
  \centering
  \includegraphics[width=\linewidth]{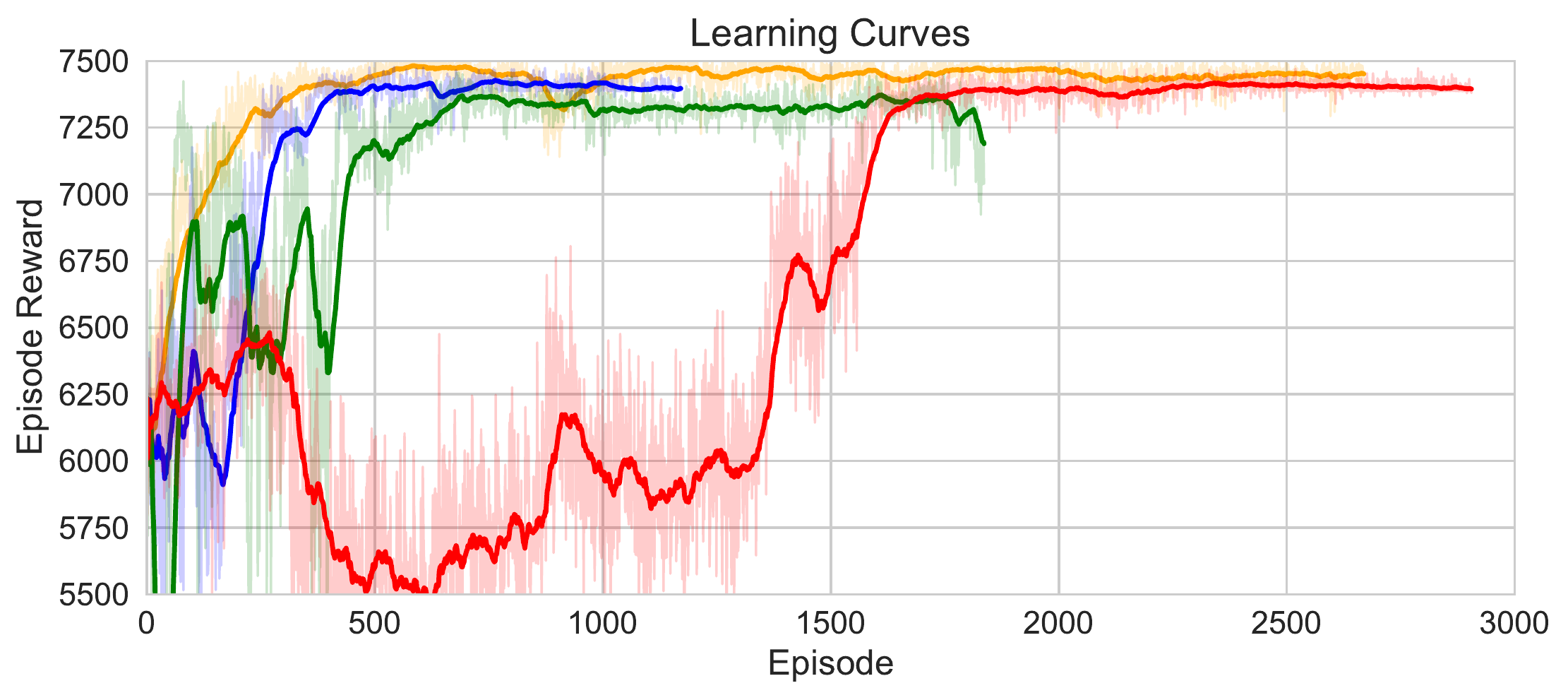}
  \caption{Smoothed learning curves of four agents, where the episode reward is the collected Fisher information.}
  \label{fig:RL_curve2}
\end{figure}

We also take a more detailed look at the progress of training and how the reward function guides the agents to reach their goals.
Figure ~\ref{fig:RL_curve2} shows smoothed learning curves of four sampled agents, trained with different hyperparameters. 
The x-axis shows the number of training episodes experienced so far, and the y-axis shows the corresponding reward achieved in each episode.
The original curves are in shaded colors, while the smoothed versions are on top in solid colors.
After sufficient episodes, all four agents reach near-optimal policies that achieve rewards close to the oracle solution.

\section{Discussion}\label{sec:dicussion}

The results show that using the mean Fisher information as a reward function is effective in learning desirable policies. %
To investigate the effectiveness and constraints of using this reward function, we can take a close look at its definition in (\ref{eq:fi}). %
Most important for IoT in a setting where taking samples is energy-intensive, the reward function does not depend on unobserved samples, only on the confidence of the resulting model over a period of time. %
Because model confidence can only come about when measurements taken at similar moments have low variance, we assume the model is close to correct when it is confident. This assumption may occasionally be unfounded in outlier moments, but measuring and hoping for an outlier which is, by definition, unlikely to occur, is not intelligent or desirable behavior.  %

The main contribution of our paper regards the development of the frugal policy, which is encoded as a neural network. This network can be easily implemented on IoT hardware. %
For example, one exemplar of a good performing policy was a network with 4 hidden layers with 32 neurons and ReLU activation functions. %
We have implemented this network on an Arduino NANO 33 BLE Sense, where it only requires 5$\pm$1 ms to evaluate. %
Compared with the estimated execution time to measure, aggregate, calculate and send noise data, this overhead is acceptable. %
The compiled flash memory consumption was 115 KB (11\,\%) and the RAM Memory was ~50 KB (20\,\%).
With regards to the GP, learning involves an inversion of the kernel matrix, which has an asymptotic complexity of $\mathcal{O}(n^3)$, where $n$ is the number of training examples. 
However, the GP can be trained off-device before deployment, so it doesn't have to store its history of measurements. 
Additionally, sparse online GP can be exploited to train incrementally without retaining the entire model when new data are observed ~\cite{csato2002sparse}. %
Furthermore, there are many model approximations or approximate GP inference with cheaper complexity that can be utilized such as Nystrom approximation ~\cite{williams2001using}, FITC approximation ~\cite{bui2017unifying}, or variational sparse GP ~\cite{titsias2009variational}. %

In general, finding good values for hyperparameters can be challenging during training. %
In our setting, we first required suitable types and parameters for the kernels of the GP. %
We found these through optimization by maximizing the likelihood of the data, using the GPy framework~\cite{gpy2014}. %
Secondly, we had to dimension the environment's parameters, such as energy capacity, prediction horizon, range and resolution for states and actions. %
For that, we used an iterative process of tuning with randomized values. %
Finally, learning an RL policy can be unstable due to the algorithm's sensitivity to hyperparameters and their intricate interplay. %
Hence only a subset of the policies performs well, and not all of them are useful. %
This, however, is not a problem since the training takes place on a device management server, and we can choose the best performing policy before deployment. %

\section{Conclusion}\label{sec:conclusion}

In this work, we have introduced a novel reward function for learning adaptive sensing policies with deep reinforcement learning. %
This reward function is based on the mean Fisher information value of a probabilistic model of an environmental phenomenon. 
The model is fitted with observations selected by the learning agent.  %
Using a case study of workplace noise monitoring, we demonstrated that this reward function led to a learned sampling policy that outperforms a uniform strategy and is close to a near-optimal oracle solution. %
Our results indicate that using this information-based reward function along with policies approximated by neural networks can achieve more generalization and autonomy in IoT applications. %
Finally, we discussed the implementations and the constraints of the proposed framework. %
While the presented work uses a case study with a temporally-varying phenomenon, it can be applied in a wide range of adaptive sensing applications. %
Overall, this approach leads to an increased level of autonomy in IoT by reducing manual design effort through information-driven behavior learning. %

\section*{Acknowledgements}
We would like to thank Finn Julius Stephansen-Smith and Amund Askeland for their implementation of the neural network.

\end{document}